\documentclass[sigconf]{acmart}

\usepackage[linesnumbered,ruled,vlined,algo2e]{algorithm2e}
\AtBeginDocument{%
  }

\setcopyright{acmlicensed}
\copyrightyear{2026}
\acmYear{2026}
\acmDOI{10.1145/3773078.3831866}
\acmConference[RecSys '26]{20th ACM Conference on Recommender Systems}{September 28 -- October 2, 2026}{Minneapolis, MN, USA}
\acmISBN{979-8-4007-2284-4/2026/09}

\begin{document}

\title[MPZCH: Zero Collision Hash for Large-Scale Recommenders]{Multi-Probe Zero Collision Hash (MPZCH): Mitigating Embedding Collisions and Enhancing Model Freshness in Large-Scale Recommenders}

\author{Ziliang Zhao}
\affiliation{%
  \institution{Meta Platforms}
  \city{Bellevue}
  \state{WA}
  \country{USA}}
\email{zlzhao@meta.com}
\orcid{0000-0002-3302-4629}

\author{Bi Xue}
\affiliation{%
  \institution{Meta Platforms}
  \city{Menlo Park}
  \state{CA}
  \country{USA}}
\email{bixue@meta.com}
\orcid{0009-0009-1917-3087}

\author{Emma Lin}
\affiliation{%
  \institution{Meta Platforms}
  \city{Bellevue}
  \state{WA}
  \country{USA}}
\email{line@meta.com}
\orcid{0009-0004-2633-8496}

\author{Tianqi Lu}
\affiliation{%
  \institution{Meta Platforms}
  \city{Bellevue}
  \state{WA}
  \country{USA}}
\email{tialu@meta.com}
\orcid{0009-0000-4634-9927}

\author{Mengjiao Zhou}
\affiliation{%
  \institution{Meta Platforms}
  \city{Bellevue}
  \state{WA}
  \country{USA}}
\email{mengjiao@meta.com}
\orcid{0009-0000-0755-4733}

\author{Kaustubh Vartak}
\affiliation{%
  \institution{Meta Platforms}
  \city{New York}
  \state{NY}
  \country{USA}}
\email{kaus@meta.com}
\orcid{0009-0005-0937-509X}

\author{Shakhzod Ali-Zade}
\affiliation{%
  \institution{Meta Platforms}
  \city{Vancouver}
  \state{BC}
  \country{Canada}}
\email{shakhzod@meta.com}
\orcid{0009-0000-6999-3408}

\author{Tao Li}
\affiliation{%
  \institution{Meta Platforms}
  \city{Bellevue}
  \state{WA}
  \country{USA}}
\email{ltao@meta.com}
\orcid{0009-0005-3572-8354}

\author{Bin Kuang}
\affiliation{%
  \institution{Meta Platforms}
  \city{Bellevue}
  \state{WA}
  \country{USA}}
\email{binkuang@meta.com}
\orcid{0000-0003-2342-2268}

\author{Rui Jian}
\affiliation{%
  \institution{Meta Platforms}
  \city{Menlo Park}
  \state{CA}
  \country{USA}}
\email{rjian@meta.com}
\orcid{0009-0007-9379-4442}

\author{Bin Wen}
\affiliation{%
  \institution{OpenAI}
  \city{San Francisco}
  \state{CA}
  \country{USA}}
\email{bwen@openai.com}
\orcid{0009-0001-8403-6080}

\author{Dennis van der Staay}
\affiliation{%
  \institution{Periodic Labs}
  \city{Menlo Park}
  \state{CA}
  \country{USA}}
\email{dennis@periodiclabs.ai}
\orcid{0009-0006-8254-4977}

\author{Yixin Bao}
\affiliation{%
  \institution{Meta Platforms}
  \city{Seattle}
  \state{WA}
  \country{USA}}
\email{yixinb@meta.com}
\orcid{0000-0002-6921-2154}

\author{Xiujin Li}
\affiliation{%
  \institution{Meta Platforms}
  \city{Bellevue}
  \state{WA}
  \country{USA}}
\email{xiujinl@meta.com}
\orcid{0009-0004-8376-6067}

\author{Chao Deng}
\affiliation{%
  \institution{Meta Platforms}
  \city{Menlo Park}
  \state{CA}
  \country{USA}}
\email{chaodeng@meta.com}
\orcid{0009-0002-3668-8102}

\author{Henry Wei}
\affiliation{%
  \institution{Meta Platforms}
  \city{Bellevue}
  \state{WA}
  \country{USA}}
\email{hongrenwei@meta.com}
\orcid{0009-0003-4606-0500}

\author{Songbin Liu}
\affiliation{%
  \institution{Meta Platforms}
  \city{New York}
  \state{NY}
  \country{USA}}
\email{songbin@meta.com}
\orcid{0009-0008-1699-8975}

\author{Qifan Wang}
\affiliation{%
  \institution{Meta Platforms}
  \city{Menlo Park}
  \state{CA}
  \country{USA}}
\email{wqfcr@meta.com}
\orcid{0000-0002-7570-5756}

\author{Kai Ren}
\affiliation{%
  \institution{Meta Platforms}
  \city{Menlo Park}
  \state{CA}
  \country{USA}}
\email{kren@meta.com}
\orcid{0009-0009-4183-9172}

\renewcommand{\shortauthors}{Zhao et al.}

\begin{abstract}
  Embedding tables are critical components of large-scale recommendation systems, facilitating the efficient mapping of high-cardinality categorical features into dense vector representations. However, as the volume of unique IDs expands, traditional hash-based indexing methods suffer from collisions that degrade model performance and personalization quality. We present Multi-Probe Zero Collision Hash (MPZCH), a novel indexing mechanism based on linear probing that effectively mitigates embedding collisions. With reasonable table sizing, it often eliminates these collisions entirely while maintaining production-scale efficiency. MPZCH utilizes auxiliary tensors and high-performance CUDA kernels to implement configurable probing and active eviction policies. By retiring obsolete IDs and resetting reassigned slots, MPZCH prevents the stale embedding inheritance typical of hash-based methods, ensuring new features learn effectively from scratch. Despite its collision-mitigation overhead, the system maintains training QPS and inference latency comparable to existing methods. Rigorous online experiments demonstrate that MPZCH achieves zero collisions for user embeddings and significantly improves item embedding freshness and quality. The solution has been released within the open-source TorchRec library for the broader community.
\end{abstract}



\maketitle

\section{Introduction}
Recommendation systems are foundational to modern digital platforms, powering personalized experiences across social media, e-commerce, online advertising, and so forth \cite{covington2016deep, zhang2019deep}. These systems leverage vast amounts of user and item data to predict preferences and deliver relevant content, products, or connections. At the core of many state-of-the-art recommendation models lies the embedding table—a structure that maps high-cardinality categorical features (such as user IDs, video IDs, interest IDs) into dense, low-dimensional vector representations.

In large-scale recommendation models, embedding tables can span tens of billions of rows, each representing a unique user or item. These massive tables are essential for capturing the full diversity and granularity of interactions \cite{pi2020practice} because they:
\begin{itemize}
\item Represent a massive and growing universe of users and items.
\item Capture nuanced relationships and similarities between entities.
\item Enable real-time, scalable personalization.
\end{itemize}

Despite their importance, embedding tables present unique challenges. Given the enormous, ever-growing number of IDs (especially item IDs) relative to memory constraints, it is often infeasible to allocate a unique row for every possible ID. In practice, the "hashing trick" \cite{weinberger2009feature} is typically used to map unique IDs to indices within a fixed-size embedding table. However, this approach introduces embedding collisions: when two or more distinct IDs are hashed to the same index, they are forced to share the same embedding vector. This phenomenon has been shown to degrade model performance by conflating unrelated users or items, thereby reducing the quality of the final recommendations \cite{liu2022monolith, zhang2020frequency}.

Beyond simple collisions, the \textit{freshness} of the embedding table poses a significant, often overlooked challenge. In dynamic environments, the distribution of active IDs shifts constantly as new items appear and old ones become obsolete. Standard hash-based methods lack a mechanism to explicitly retire obsolete IDs; consequently, when a new ID collides with a slot occupied by a stale ID, it effectively "inherits" a pre-trained embedding that is wholly unrelated to its true semantics. This negative transfer hinders the model's ability to learn accurate representations for new items, forcing them to unlearn noise before converging on useful signals.

While increasing the size of the embedding table can reduce the probability of collisions, it cannot eliminate them entirely—especially as the number of unique IDs continues to grow, since embedding tables cannot be expanded indefinitely due to memory limitations. In fact, even if the table is large enough to contain all possible IDs, the hash algorithm could still result in significant collisions \cite{shi2020compositional}. Therefore, managing hash collisions and ensuring embedding freshness remain fundamental challenges.

In this paper, we focus on the problem of embedding collisions and staleness in large-scale recommendation models. We propose a novel, high-performance solution that mitigates collisions and actively manages ID lifecycles, thereby improving the reliability and effectiveness of recommendations at scale.

\section{Related Work}
The challenge of hash collisions in embedding tables remains a critical bottleneck for large-scale recommendation systems, where mapping billions of unique IDs to dense vectors requires balancing model quality with memory constraints. To address this, research has evolved from strict collision avoidance to probabilistic and semantic adaptation.

A primary direction is the pursuit of collisionless hashing to ensure high fidelity for important entities. Bytedance’s Monolith \cite{liu2022monolith} exemplifies this by employing Cuckoo Hashing to guarantee unique slots for high-frequency IDs, supported by dynamic expiration to manage table growth. While effective, the memory overhead scales linearly with feature cardinality. To address this capacity challenge without relying on hashing, Mixed Dimension Embeddings \cite{ginart2021mixed} abandon fixed-width tables entirely, optimizing the information-per-bit ratio by assigning variable embedding dimensions based on item popularity.

To reduce collision probability without the infinite memory required for collisionless tables, various hashing techniques have been explored. Twitter \cite{yang2020twitter} utilizes double hashing to statistically minimize conflicts, while Multiplexed Embeddings \cite{huan2023multiplexed} concatenate outputs from multiple hash functions to increase the effective embedding space. Frequency-Based Double Hashing \cite{zhang2020frequency} refines these concepts into a hybrid approach: it applies strict uniqueness to "head" items (similar to Monolith) but utilizes double hashing specifically for "tail" items, balancing precision for popular entities with memory efficiency for the long tail.

Distinct from standard hashing, recent work focuses on compositional uniqueness. The "QR trick" \cite{shi2020compositional} maps IDs to complementary partitions (quotient and remainder), combining two smaller lookups to form a unique vector. Similarly, TT-Rec \cite{yin2021ttrec} applies Tensor Train decomposition to factorize embeddings into compact multidimensional tensors. These methods achieve virtual uniqueness with significantly fewer parameters, though they often trade memory savings for increased inference latency due to complex reconstruction logic.

Rather than treating collisions as purely statistical errors, some methods integrate them into the learning process. Google’s Deep Hash Embedding \cite{kang2020dhe} uses neural layers to "disentangle" shared representations, replacing simple lookups with deep transformations. Taking this further, "Learning to Collide" \cite{ghaemmaghami2022collide} and Semantic IDs \cite{kulkarni2024semantic} leverage semantic clustering to ensure that collisions occur only between similar items. While this benefits tail items and stability, it introduces dependency on preprocessing and can lead to over-smoothing of distinct features.

Finally, highly aggressive compression techniques prioritize cache locality and hardware efficiency. Unified Embedding \cite{zhang2023unified} multiplexes multiple features into a single table, reducing the memory footprint at the risk of inter-feature interference. ROBE \cite{desai2022robe} pushes this extreme by using a single circular buffer with random block offsets, matching the performance of full tables with orders of magnitude less memory. These approaches maximize system scalability but require the model to be highly robust against the noise introduced by hash collisions.

\section{Methodology}
The proposed algorithm is implemented as a high-performance CUDA kernel that handles insertion, lookup, and eviction operations for tens of thousands of IDs in parallel. However, achieving collision mitigation introduces overhead compared to traditional hash-based approaches. Specifically, MPZCH requires storing additional state to track row occupancy and freshness, ensuring optimal utilization of available slots.

\subsection{Auxiliary Tensors}
To manage collision resolution, MPZCH maintains two key auxiliary tensors—the \textit{identities} tensor and the \textit{metadata} tensor—which operate in conjunction with the embedding table. The identities tensor (initialized to -1) records the specific ID occupying each slot, allowing the kernel to strictly distinguish between occupied slots (identity == ID) and available slots (identity == -1). During training, if a lookup fails (ID is missing), the kernel attempts to allocate a dedicated slot for the new ID by either locating an empty row or evicting an expired entry. The optional metadata tensor stores auxiliary information, such as time-to-live (TTL) values, to guide eviction decisions.

\subsection{Linear Probing Mechanism}
The core of MPZCH is a linear probing mechanism. The simplified process for handling a single ID is illustrated in Algorithm \ref{alg:mpzch_algo} (omitting kernel dispatch and synchronization details). During model training, each ID is mapped to an initial slot via a hash function, ensuring a uniform distribution across the embedding table within each shard.

The probing process executes a two-pass scan starting from the initial slot: \begin{enumerate} \item \textbf{Pass 1 (Discovery):} The kernel scans the probe range solely to determine if the ID is already present. It does not perform any write operations during this phase. \item \textbf{Pass 2 (Action):} The kernel iterates through the range again to perform the necessary operation based on the result of the first pass: \begin{itemize} \item \textit{Update:} If the ID was discovered in Pass 1, the kernel updates the metadata (e.g., timestamp) of the existing slot. \item \textit{Allocation:} If the ID was not found in Pass 1, the kernel attempts to allocate a slot. This results in one of three outcomes: (1) \textit{Insertion} into an empty slot; (2) \textit{Eviction} of an expired ID (if enabled, see \ref{sec:eviction_management}); or (3) \textit{Collision}, where the kernel defaults to returning the initial slot if no empty or evitable slot is found. \end{itemize} \end{enumerate}

The two-pass design mitigates potential consistency issues inherent to sequential linear probing. Without this separation, an incoming ID might evict a stale slot early in the probe range, even if that same ID already exists further down the range (e.g., allocated during a previous step before the earlier slot expired). By separating discovery from action, MPZCH ensures that existing entries are always prioritized over new allocations, preventing duplicate storage of the same ID.

\SetKwFunction{FLookup}{Lookup}
\SetKwBlock{Begin}{}{end}

\begin{algorithm2e*}[t]
\caption{MPZCH Kernel: Look Up a Single ID}
  \Description{MPZCH Kernel: Look Up a Single ID}
\label{alg:mpzch_algo}

\KwIn{Query ID $id$, Metadata $meta$ (e.g., Expiration Threshold)}
\KwOut{Tuple $(slot, evicted)$ - mapped slot index and eviction flag}

\textbf{Global:} $\mathcal{I}$ (Identities Tensor), $\mathcal{M}$ (Metadata Tensor), $P$ (Max Probe) \\
\textbf{Constants:} $\emptyset$ (Empty Indicator, initialized to -1)
\BlankLine

\noindent\textbf{Function} \FLookup{$id$, $meta$} \Begin{
    $h \gets \textsc{Hash}(id)$ \tcp*[r]{Compute initial hash index}
    $exists \gets \textsc{False}$\;
    
    \tcp{Pass 1: Check if ID already exists in probe range}
    \For{$i \gets 0$ \textbf{to} $P - 1$}{
        $slot \gets h + i$\;
        \If{$\mathcal{I}[slot] = id$}{
            $exists \gets \textsc{True}$\;
            \textbf{break} \tcp*[r]{Found existing ID, stop scanning}
        }
    }
    
    \tcp{Pass 2: Update, Insert, or Evict}
    \For{$i \gets 0$ \textbf{to} $P - 1$}{
        $slot \gets h + i$\;
        $id_{curr} \gets \mathcal{I}[slot]$\;
        $meta_{curr} \gets \mathcal{M}[slot]$\;
        
        \tcp{Case 1: Found the ID (Update) or an Empty Slot (Insert)}
        \If{$id_{curr} = id$ \textbf{or} $id_{curr} = \emptyset$}{
            $\mathcal{I}[slot] \gets id$\;
            $\mathcal{M}[slot] \gets meta$ \tcp*[r]{Refresh TTL}
            \KwRet $(slot, \textsc{False})$\;
        }
        
        \tcp{Case 2: Eviction Candidate}
        \If{\textbf{not} $exists$ \textbf{and} $meta > meta_{curr}$}{
            $\mathcal{I}[slot] \gets id$\;
            $\mathcal{M}[slot] \gets meta$\;
            \KwRet $(slot, \textsc{True})$ \tcp*[r]{Signal eviction}
        }
    }
    
    \tcp{Case 3: Collision Fallback}
    $\mathcal{M}[h] \gets meta$\;
    \KwRet $(h, \textsc{False})$\;
}
\end{algorithm2e*}

\subsection{Kernel Execution}
\begin{figure*}
  \centering
  \includegraphics[width=\linewidth]{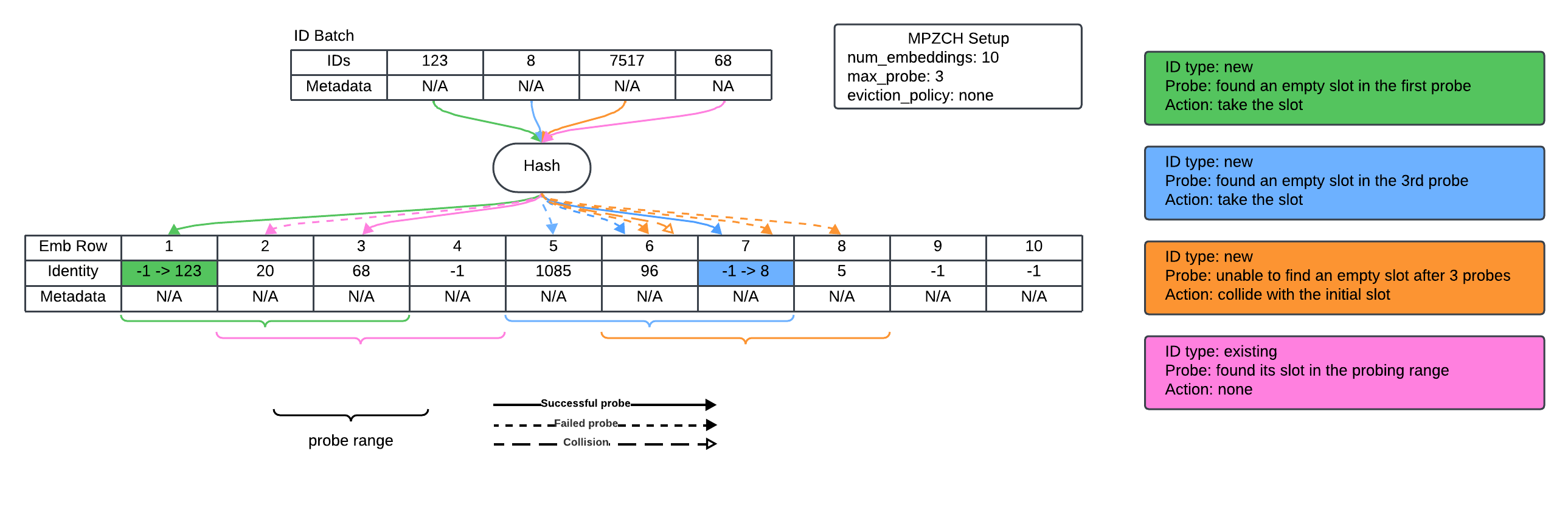}
  \caption{A simplified example of ID insertion, lookup, and collision handling.}
  \Description{A simplified example of ID insertion, lookup, and collision handling.}
  \label{fig:mpzch_basic}
\end{figure*}

During training, the data pipeline provides a batch of IDs along with the shard’s identities and metadata tensors. The kernel distributes these IDs across GPU threads to perform parallel lookup, insertion, and eviction operations. Figure \ref{fig:mpzch_basic} illustrates the lookup and insertion process with eviction disabled, demonstrating how linear probing navigates occupied and empty slots.

Table \ref{tab:collision_rate} demonstrates the relationship between table size, the \texttt{max\_probe} parameter, and collision rates using 150 million unique user IDs. We compare MPZCH against the standard hash-based method (\texttt{torcharrow.functional.sigrid\_hash}), the default algorithm for most models at Meta. Key findings include:

\begin{itemize} \item While larger tables generally reduce collisions, hash-based methods continue to suffer significant collision rates even when the table size far exceeds ID cardinality. \item MPZCH significantly reduces collisions even with limited table sizes. With a table size of 200 million (slightly larger than the 150M cardinality) and a probe depth of 256, MPZCH achieves zero collisions, whereas Sigrid hash incurs a nearly 30\% collision rate. Increasing the table size to 300 million allows MPZCH to achieve zero collisions with a reduced probe depth of 64. \end{itemize}

\begin{table*}[t]
  \caption{Collision rate comparison between Sigrid Hash and MPZCH using 150 million unique IDs. The Capacity Ratio represents the degree of over-provisioning, calculated as the embedding table size divided by the ID cardinality ($N_{table} / N_{ids}$). The results demonstrate that MPZCH achieves zero collisions with adequate capacity and probe depth, whereas the baseline Sigrid Hash retains significant collisions even when the table size is more than triple the ID cardinality.}
  \Description{Collision rate comparison between Sigrid Hash and MPZCH using 150 million unique IDs. The Capacity Ratio represents the degree of over-provisioning, calculated as the embedding table size divided by the ID cardinality ($N_{table} / N_{ids}$). The results demonstrate that MPZCH achieves zero collisions with adequate capacity and probe depth, whereas the baseline Sigrid Hash retains significant collisions even when the table size is more than triple the ID cardinality.}
  \label{tab:collision_rate}
  \centering
  \resizebox{\textwidth}{!}{
  \begin{tabular}{cc||c|ccccccc}
    \toprule
    \textbf{Table Size} & \textbf{Capacity} & \textbf{Sigrid Hash} & \multicolumn{7}{c}{\textbf{MPZCH Collision Rate (\%) at max\_probe = $P$}} \\
    (Millions) & Ratio & \textbf{Collision} & $P=8$ & $P=16$ & $P=32$ & $P=64$ & $P=128$ & $P=256$ & $P=512$ \\
    \midrule
    100 & 0.67x & 48.2080\% & 34.0631\% & 33.4269\% & 33.3363\% & 33.3333\% & 33.3333\% & 33.3333\% & 33.3333\% \\
    150 & 1.00x & 36.7917\% & 12.0940\% & 8.4059\% & 5.8717\% & 4.1186\% & 2.8981\% & 2.0430\% & 1.4411\% \\
    200 & 1.33x & 29.6472\% & 3.8475\% & 1.3054\% & 0.2875\% & 0.0299\% & 0.0008\% & \textbf{0.0000\%} & \textbf{0.0000\%} \\
    250 & 1.67x & 24.8028\% & 1.2974\% & 0.1967\% & 0.0105\% & 0.0001\% & \textbf{0.0000\%} & \textbf{0.0000\%} & \textbf{0.0000\%} \\
    300 & 2.00x & 21.3082\% & 0.4791\% & 0.0332\% & 0.0004\% & \textbf{0.0000\%} & \textbf{0.0000\%} & \textbf{0.0000\%} & \textbf{0.0000\%} \\
    350 & 2.33x & 18.6686\% & 0.1957\% & 0.0064\% & \textbf{0.0000\%} & \textbf{0.0000\%} & \textbf{0.0000\%} & \textbf{0.0000\%} & \textbf{0.0000\%} \\
    400 & 2.67x & 16.6069\% & 0.0864\% & 0.0014\% & \textbf{0.0000\%} & \textbf{0.0000\%} & \textbf{0.0000\%} & \textbf{0.0000\%} & \textbf{0.0000\%} \\
    450 & 3.00x & 14.9618\% & 0.0407\% & 0.0003\% & \textbf{0.0000\%} & \textbf{0.0000\%} & \textbf{0.0000\%} & \textbf{0.0000\%} & \textbf{0.0000\%} \\
    500 & 3.33x & 13.6052\% & 0.0206\% & 0.0001\% & \textbf{0.0000\%} & \textbf{0.0000\%} & \textbf{0.0000\%} & \textbf{0.0000\%} & \textbf{0.0000\%} \\
    \bottomrule
  \end{tabular}%
  }
\end{table*}

\subsection{Eviction Management}
\label{sec:eviction_management}
\begin{figure*}
  \centering
  \includegraphics[width=\linewidth]{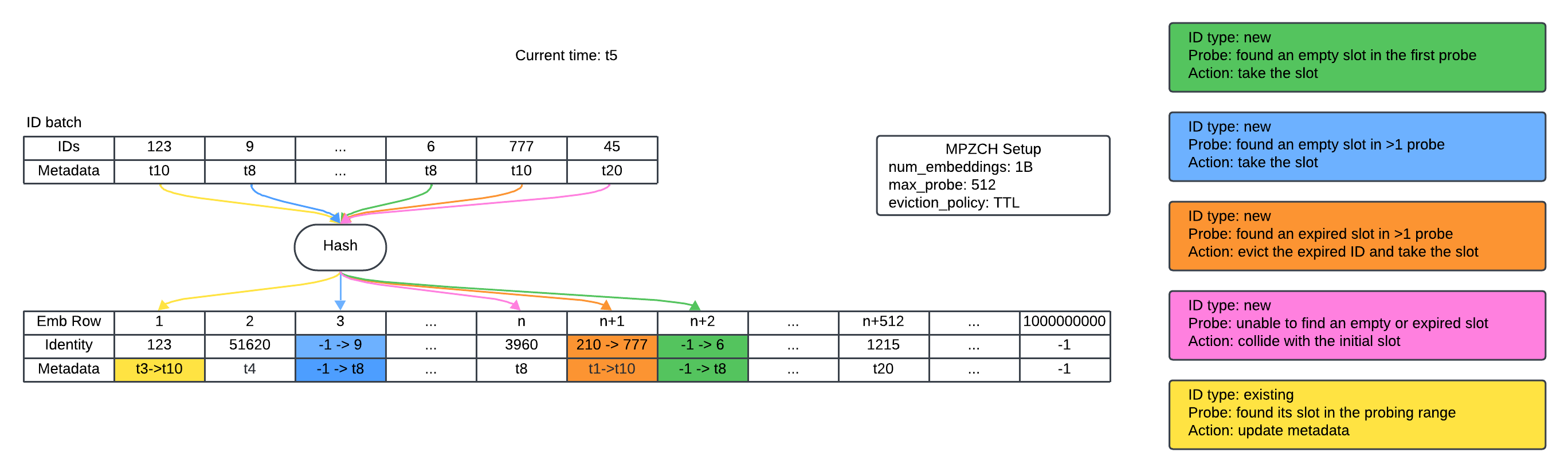}
  \caption{An example of kernel execution with TTL eviction policy. Probing details are omitted for clarity, with arrows pointing to the final result slots.}
  \Description{An example of kernel execution with TTL eviction policy. Probing details are omitted for clarity, with arrows pointing to the final result slots.}
  \label{fig:mpzch_eviction}
\end{figure*}

MPZCH offers configurable eviction strategies to address diverse application requirements. The standard mechanism utilizes a Time-To-Live (TTL) policy. A TTL value is computed based on the current training timestamp and a user-defined duration (e.g., three days); this value is updated in the metadata tensor whenever an ID is processed. Any slot with a metadata value lower than the current timestamp is deemed expired. MPZCH employs a \textit{lazy eviction} strategy, meaning eviction is triggered only when a new ID requires allocation. As illustrated in Figure \ref{fig:mpzch_eviction}, if a target ID is not found, the algorithm reclaims the first expired slot encountered in the probe range.

This mechanism also supports per-feature TTL configurations for embedding tables that manage multiple distinct features. It allows for granular control over retention strategies: critical features can be assigned longer retention periods to preserve high-value data, while less significant features are evicted more aggressively to optimize capacity.

Crucially, upon eviction, both the embedding weights and the optimizer state (e.g., momentum) are reset. Unlike hash-based methods where a new ID inherits the pre-trained embedding of a colliding ID, MPZCH ensures that the new item learns from scratch. It prevents "negative transfer" from unrelated semantic data, allowing the model to learn accurate representations more effectively. This reset mechanism is the primary driver of the freshness improvements observed in AB testing.

\begin{figure*}
  \centering
  \includegraphics[width=\linewidth]{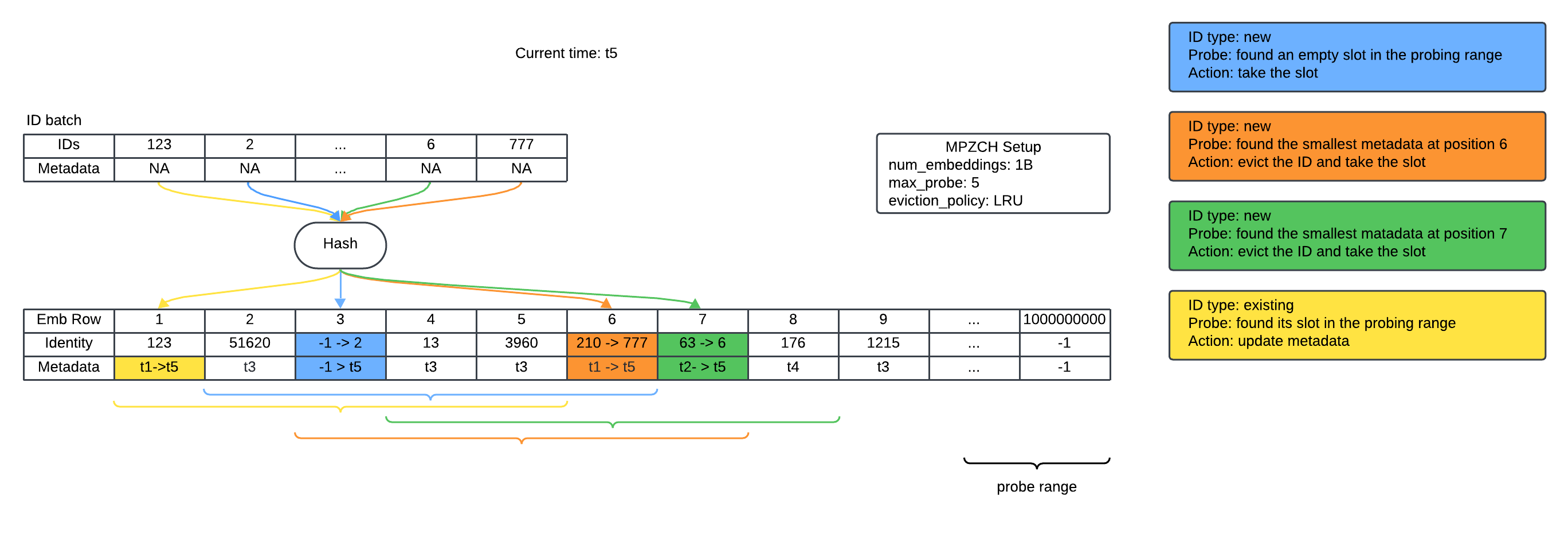}
  \caption{An example of kernel execution with LRU eviction policy. Probing details are omitted for clarity, with arrows pointing to the final result slots.}
  \Description{An example of kernel execution with LRU eviction policy. Probing details are omitted for clarity, with arrows pointing to the final result slots.}
  \label{fig:mpzch_lru}
\end{figure*}

We also support an LRU (Least Recently Used) policy (Figure \ref{fig:mpzch_lru}), where the algorithm scans the probe range to identify and evict the entry with the oldest timestamp, thereby preserving recently seen IDs.

\subsection{Performance Benchmark}
Table \ref{tab:kernel_benchmark} presents timing measurements across different memory configurations (HBM/CUDA, UVM\_Managed, CPU/DRAM) in both batched and non-batched modes. Batched operations on HBM/CUDA achieve optimal performance (approximately 0.8–0.9 ms), whereas non-batched CPU/DRAM operations exhibit significantly higher latency (95–100 ms), underscoring the necessity of GPU acceleration.

Furthermore, the data indicates that the \texttt{max\_probe} depth has a negligible impact on latency during both training and inference. Consequently, production deployments typically utilize a sufficiently large probe depth to minimize collisions without incurring performance penalties.

\begin{table*}[t]
  \caption{MPZCH kernel latency benchmark (in milliseconds). The results demonstrate that MPZCH performance is highly efficient on GPU and remains consistent across varying probe depths, indicating that the linear probing mechanism incurs negligible overhead even at higher search ranges.}
  \Description{MPZCH kernel latency benchmark (in milliseconds). The results demonstrate that MPZCH performance is highly efficient on GPU and remains consistent across varying probe depths, indicating that the linear probing mechanism incurs negligible overhead even at higher search ranges.}
  \label{tab:kernel_benchmark}
  \centering
  \resizebox{0.8\textwidth}{!}{%
  \begin{tabular}{l|ccccccc}
    \toprule
    \textbf{Hardware \& Execution Mode} & \multicolumn{7}{c}{\textbf{Latency (ms) at max\_probe = $P$}} \\
    \cmidrule(l){2-8}
     & $P=8$ & $P=16$ & $P=32$ & $P=64$ & $P=128$ & $P=256$ & $P=512$ \\
    \midrule
    \multicolumn{8}{l}{\textit{HBM / CUDA (GPU Native)}} \\
    \hspace{3mm} Batched & 0.90 & 0.80 & 0.90 & 0.80 & 0.90 & 0.80 & 0.90 \\
    \hspace{3mm} Non-batched & 14.60 & 14.50 & 14.70 & 14.50 & 14.50 & 14.50 & 14.60 \\
    \midrule
    \multicolumn{8}{l}{\textit{UVM Managed (Unified Virtual Memory)}} \\
    \hspace{3mm} Batched & 4.40 & 4.40 & 4.40 & 4.40 & 4.40 & 4.40 & 4.40 \\
    \hspace{3mm} Non-batched & 14.60 & 14.50 & 14.40 & 14.40 & 14.50 & 14.50 & 14.60 \\
    \midrule
    \multicolumn{8}{l}{\textit{CPU / DRAM}} \\
    \hspace{3mm} Batched & 8.90 & 7.10 & 6.10 & 7.20 & 6.40 & 6.40 & 6.80 \\
    \hspace{3mm} Non-batched & 102.70 & 95.10 & 97.00 & 96.90 & 97.50 & 94.20 & 98.40 \\
    \bottomrule
  \end{tabular}%
  }
\end{table*}

\subsection{Integration with TorchRec}
TorchRec \cite{ivchenko2022torchrec}, open-sourced alongside PyTorch, is the standard for large-scale distributed training at Meta. Our architectural design explicitly targets seamless integration with TorchRec’s abstractions of \texttt{EmbeddingBagCollection} and \texttt{EmbeddingCollection}, which serve as the foundation for embedding modules in a distributed setting.

\begin{figure*}
  \centering
  \includegraphics[width=0.8\textwidth]{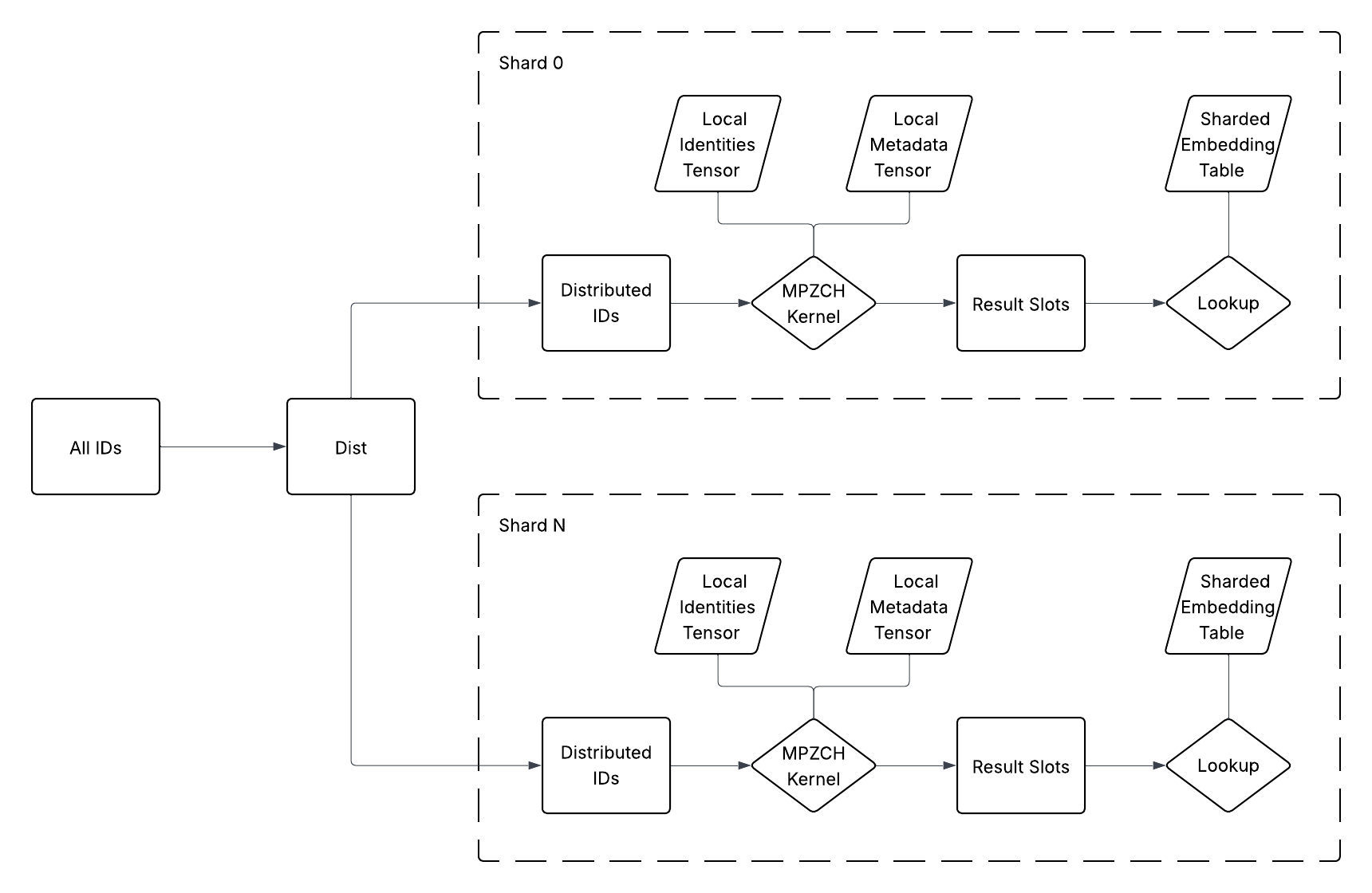}
  \caption{The MPZCH module is co-sharded with the associated embedding table, ensuring that operations are executed locally within each shard.}
  \Description{The MPZCH module is co-sharded with the associated embedding table, ensuring that operations are executed locally within each shard.}
  \label{fig:mpzch_sharding}
\end{figure*}

By strictly aligning with these established abstractions, we made a strategic design choice to leverage TorchRec’s robust infrastructure for efficient memory management and flexible sharding strategies. This approach ensures that MPZCH delivers a performant, collision-mitigating embedding table that remains compatible with existing production architectures, obviating the need to reinvent the underlying distributed systems stack. This integration necessitates strict distributed data alignment; specifically, the identity and metadata tensors must be row-wise sharded across ranks to correspond exactly with the embedding table (Figure \ref{fig:mpzch_sharding}). This configuration guarantees that all lookup, insertion, and eviction operations remain local to the shard, eliminating the need for complex inter-rank synchronization during kernel execution. Furthermore, the implementation supports input feature deduplication \cite{zhao2023recd} to mitigate training QPS regression and utilizes TorchRec’s native mechanisms for automatically resetting embedding and optimizer states upon eviction.

\subsection{Publish and Inference}
During the model publishing process, it is essential to ensure both efficiency and correctness for downstream inference. The identities tensor is published alongside the model, as it is indispensable for runtime ID lookup. During inference, no new insertions, modifications, or evictions are expected to take place. As a result, model publish freezes MPZCH's auxiliary tensors so it becomes strictly read only in the inference service, thereby providing deterministic and highly efficient embedding lookups. For the same reason, the metadata tensor—used during training to manage eviction and freshness—is omitted from the published model snapshot, because retaining this tensor only increases the model’s size without providing any benefit for inference.

\subsection{Online Training with Streaming Updates}
To improve model freshness and user engagement, critical recommendation models typically employ online training and streaming updates \cite{matam2024quickupdate}. Complementing infrequent full-snapshot updates, the system executes sparse delta publishing at minute-level intervals, propagating only modified embedding rows to the inference service. For MPZCH, maintaining strict consistency between the identities tensor and the embedding table is a fundamental requirement. To address this, the update mechanism guarantees that the corresponding entries in the identities tensor are updated in conjunction with the embedding rows, thereby preserving the integrity of lookup operations. This comprehensive support for online training ensures MPZCH-powered models remain highly responsive to dynamic, real-time updates in the ecosystem.

\section{Experimentation}
In this section, we evaluate the performance improvements yielded by MPZCH in production environments, drawing on results from large-scale A/B tests. We categorize our analysis into two distinct domains: User Embeddings, where the primary goal is collision elimination, and Item Embeddings, where the focus shifts to freshness and rapid convergence.

\subsection{MPZCH for user embeddings}
User embedding tables typically exhibit predictable growth patterns and relative stability over time. In this context, the primary objective is to guarantee a dedicated embedding representation for every unique user by achieving zero collisions. As demonstrated in our methodology, MPZCH ensures zero collisions provided the table size exceeds the cardinality of the ID set and a sufficient probe depth (e.g., 512) is employed.

We deployed MPZCH in a production Video Late-Stage Ranking model serving approximately 3 billion monthly active users. We allocated a 4 billion-row embedding table, providing a capacity buffer to accommodate future growth while ensuring a collision-free environment.

The elimination of collisions translated directly into model quality gains. As detailed in Table \ref{tab:result_ne}, the integration of MPZCH resulted in Normalized Entropy (NE) improvements in 14 out of 17 prediction tasks, with the remaining 3 tasks showing neutral results. Given the maturity of this ranking model, which has undergone years of iterative optimization, these gains are statistically significant and represent a meaningful uplift in prediction accuracy.

\begin{table}[t]
  \caption{Normalized Entropy (NE) improvements for critical tasks following the deployment of MPZCH for the user embedding table.}
  \Description{Normalized Entropy (NE) improvements for critical tasks following the deployment of MPZCH for the user embedding table.}
  \label{tab:result_ne}
  \centering
  \begin{tabular}{lc}
    \toprule
    \textbf{Prediction Task} & \textbf{NE Improvement} \\
    \midrule
    Share & 0.38\% \\
    Video View Duration (VVD) & 0.12\% \\
    Video View Percentage 100\% (VVP100) & 0.12\% \\
    Skip & 0.09\% \\
    \bottomrule
  \end{tabular}
\end{table}

\subsection{MPZCH for item embeddings}
For item embeddings (e.g., videos, posts), the challenge is not only capacity but also the rapid turnover of relevant content. MPZCH addresses this by leveraging its eviction mechanism to enhance embedding freshness and quality through two key mechanisms: \begin{itemize} \item \textbf{Capacity Optimization:} Eviction proactively removes obsolete items—typically those unlikely to resurface—thereby freeing limited table space for fresher, higher-value content. \item \textbf{Optimizer Reset:} By resetting the optimizer state (e.g., momentum) upon eviction, MPZCH ensures that new items start learning from a neutral state rather than inheriting from a stale and unrelated item. \end{itemize}

We integrated MPZCH into a key video retrieval model based on the HSTU architecture \cite{zhai2024actions}, targeting two critical embedding tables: the \textit{Post ID} table and the \textit{Owner ID} table. We explicitly tuned the TTL configurations to align with the temporal characteristics of the video landscape, ultimately selecting a 24-hour duration for Post IDs and a 72-hour duration for Owner IDs.

A/B testing demonstrated that this configuration yielded a 0.83\% more impression for new videos (posted within 48 hours), along with quality gains. To understand the mechanism behind, we conducted a qualitative analysis using t-SNE (t-distributed Stochastic Neighbor Embedding)—a popular dimensionality reduction technique especially used for visualizing high-dimensional data such as embeddings—to assess the impact of collision reduction on embedding quality.

\subsubsection{Enhanced Semantic Correlation}
\begin{figure*}
  \centering
  \includegraphics[width=0.5\textwidth]{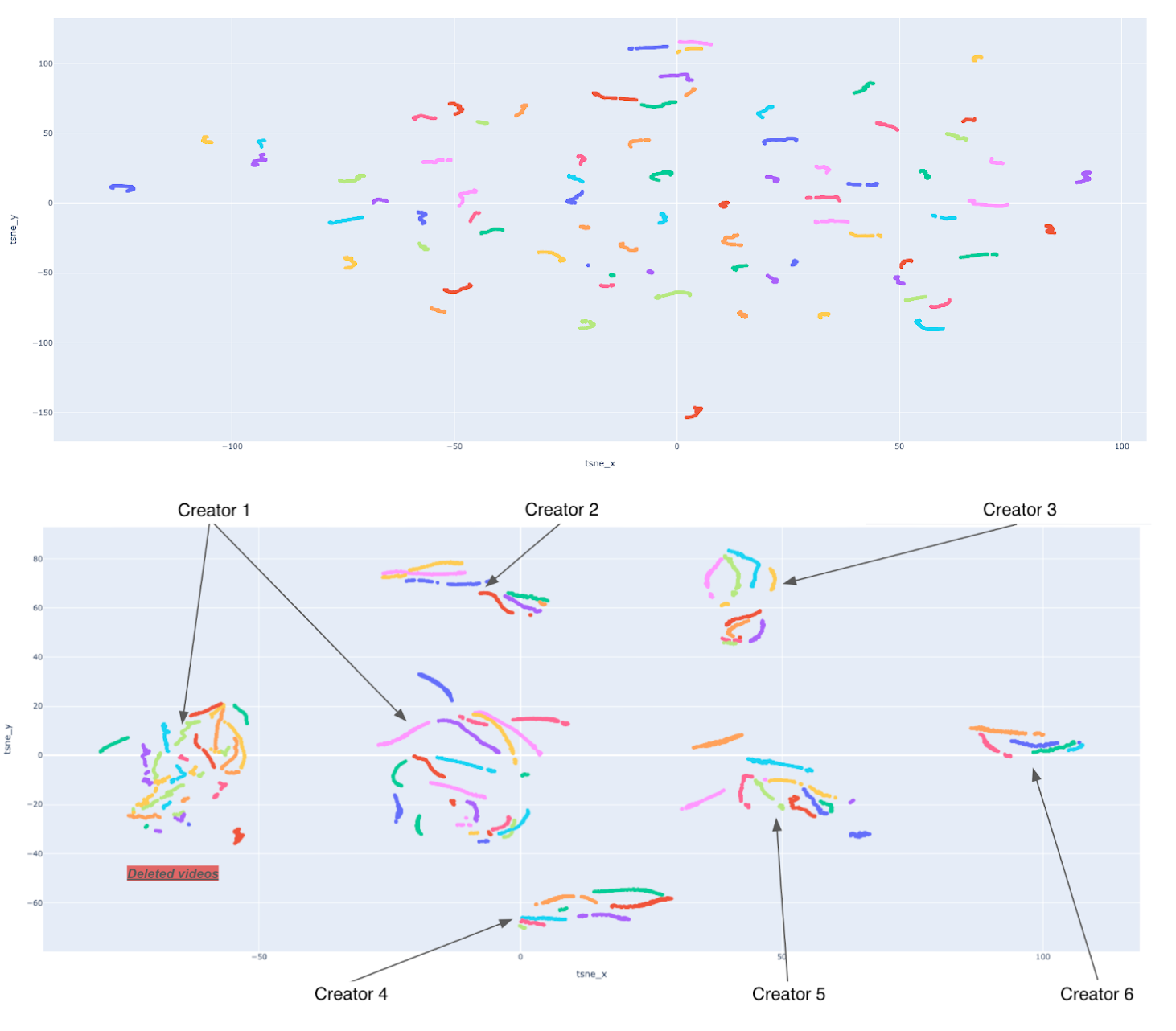}
  \caption{Top: Video embeddings from selected creators in the baseline model appear dispersed and uncorrelated. Bottom: With MPZCH, embeddings for the same creators exhibit significantly higher correlation and tighter clustering.}
  \Description{Top: Video embeddings from selected creators in the baseline model appear dispersed and uncorrelated. Bottom: With MPZCH, embeddings for the same creators exhibit significantly higher correlation and tighter clustering.}
  \label{fig:embedding_init}
\end{figure*}

We analyzed the initial embedding states of recently posted videos from selected creators (Figure \ref{fig:embedding_init}). In the baseline production model (standard hashing), the embeddings for videos from the same creator appeared stochastically distributed, indicating a lack of initial semantic coherence due to hash collisions.

In contrast, the MPZCH-enabled model exhibited significantly higher clustering among videos belonging to the same owner from the outset. We quantified this improvement by measuring the average similarity of video embeddings from the same creator; MPZCH increased this similarity metric from 25\% to 38\% (Table \ref{tab:embedding_init}).

\begin{table}[t]
  \caption{Comparison of intra-creator embedding similarity. MPZCH consistently yields higher similarity scores across all creation time windows, with the most significant gains observed for videos posted in close temporal proximity.}
  \Description{Comparison of intra-creator embedding similarity. MPZCH consistently yields higher similarity scores across all creation time windows, with the most significant gains observed for videos posted in close temporal proximity.}
  \label{tab:embedding_init}
  \centering
  \begin{tabular}{lccc}
    \toprule
    \textbf{Creation Window} & \textbf{Production} & \textbf{MPZCH} & \textbf{Relative Lift} \\
    \midrule
    Same Day & 0.66\% & 0.91\% & +38\% \\
    Same \& Next Day & 0.66\% & 0.91\% & +38\% \\
    Overall (All Time) & 0.62\% & 0.77\% & +25\% \\
    \bottomrule
  \end{tabular}
\end{table}

This increased coherence is particularly beneficial for original-content creators who maintain consistent thematic elements. By leveraging the Owner ID feature more effectively, MPZCH allows the model to rapidly establish strong, owner-correlated embeddings. This accelerates the "cold start" phase for new content, enabling creators to expand their distribution and organic reach more quickly.

\subsubsection{Stabilized Learning Trajectories}
\begin{figure*}
  \centering
  \includegraphics[width=0.5\textwidth]{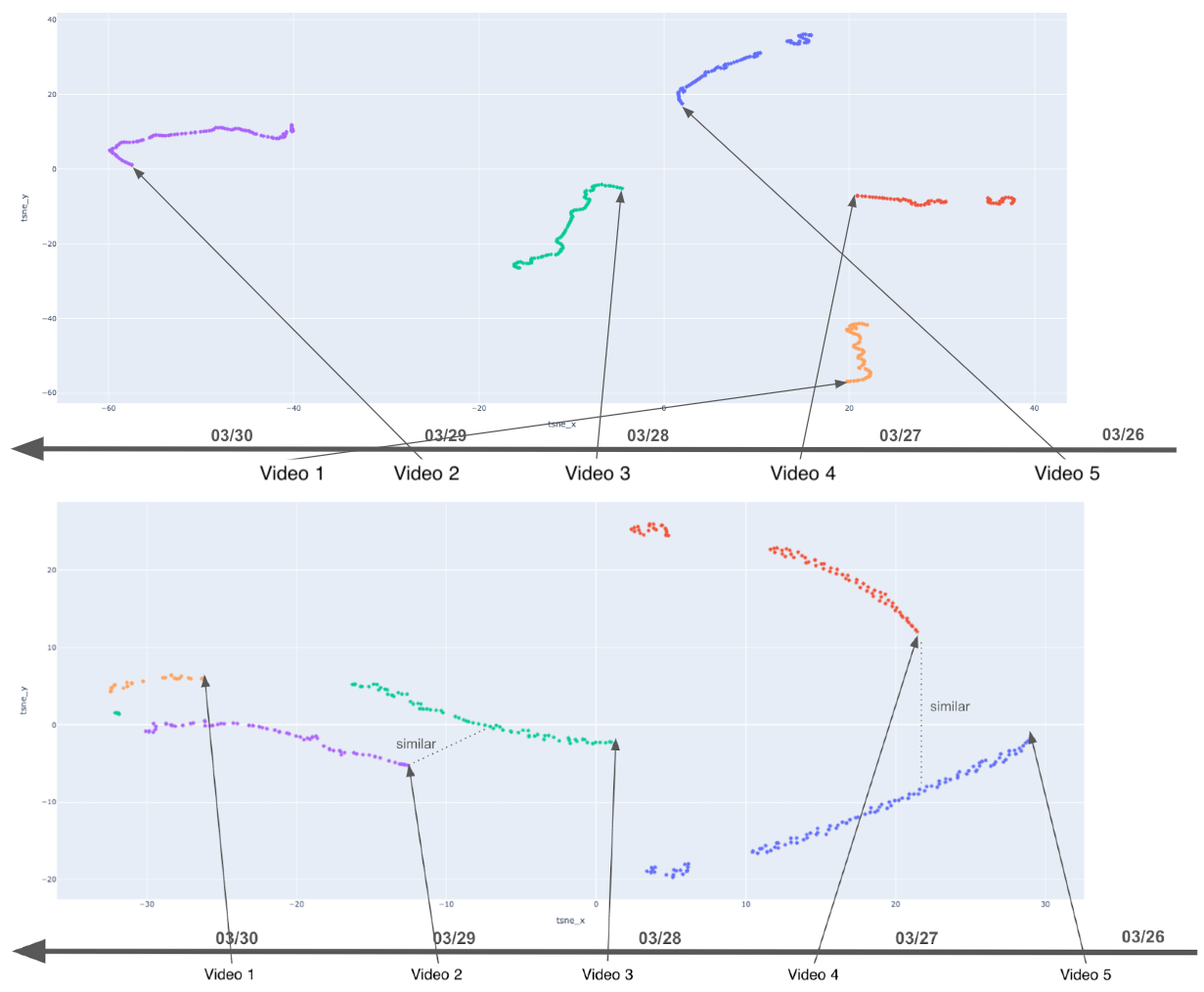}
  \caption{Top: In the absence of MPZCH, video embeddings exhibit volatile shifts attributed to hash collisions. Bottom: MPZCH enables the learning trajectory to follow a consistent, smooth path, free from collision-induced noise.}
  \Description{Top: In the absence of MPZCH, video embeddings exhibit volatile shifts attributed to hash collisions. Bottom: MPZCH enables the learning trajectory to follow a consistent, smooth path, free from collision-induced noise.}
  \label{fig:embedding_trend}
\end{figure*}

We further observed that in the baseline model, video embeddings often exhibited erratic changes across training checkpoints (Figure \ref{fig:embedding_trend}). We hypothesize that these volatile shifts are artifacts of hash collisions, where gradient updates for one item inadvertently destabilize the representation of another sharing the same slot.

With MPZCH, the learning trajectories are markedly smoother. As the model trains, the updates to video embeddings follow a consistent directional path. This stability confirms that a collision-free environment facilitates more efficient gradient descent, allowing the model to refine item representations without the interference of collision noise.

\section{Conclusion}
In this work, we address the persistent challenge of embedding collisions in large-scale recommendation models by introducing the Multi-Probe Zero Collision Hash (MPZCH) algorithm. MPZCH synergizes efficient linear probing with auxiliary tensors for precise identity management and configurable eviction policies, offering a dual-advantage solution for modern recommendation systems.

First, by guaranteeing a collision-free environment, MPZCH significantly elevates embedding quality. By eliminating the stochastic interference inherent in standard hashing, the algorithm ensures that every entity retains a dedicated, high-fidelity representation, directly translating to improved model convergence and prediction accuracy as evidenced by NE gains. Second, the introduction of effective eviction strategies fundamentally enhances embedding freshness. By proactively retiring obsolete IDs, MPZCH optimizes table utilization for high-value, temporally relevant content, enabling the model to adapt rapidly to shifting data distributions without unbounded memory growth. Last but not least, empirical evaluations across diverse production models confirm these benefits, delivering substantial improvements in model freshness and quality.

\bibliographystyle{ACM-Reference-Format}
\bibliography{main}

\end{document}